\documentclass[11pt]{article}

\usepackage{colacl}
\usepackage{graphicx}

\title{A Probabilistic Method for Analyzing Japanese Anaphora Integrating Zero Pronoun Detection and Resolution}

\author{Kazuhiro Seki$^{\,\dag}$, Atsushi Fujii$^{\,\dag\dag,\,\dag\dag\dag}$
  and Tetsuya Ishikawa$^{\,\dag\dag}$\\
  \dag National Institute of Advanced Industrial Science and Technology\\
  {\normalsize 1-1-1, Chuuou Daini Umezono, Tsukuba 305-8568, Japan}\\
  \dag\dag University of Library and Information Science \\
  {\normalsize 1-2, Kasuga, Tsukuba, 305-8550, Japan}\\
  \dag\dag\dag  CREST, Japan Science \& Technology Corporation \\
  {\normalsize\tt k.seki@aist.go.jp\ \ \ fujii@ulis.ac.jp\ \ \ ishikawa@ulis.ac.jp}}

\begin{document}
\maketitle

\begin{abstract}
  This paper proposes a method to analyze Japanese anaphora, in which
  zero pronouns (omitted obligatory cases) are used to refer to
  preceding entities (antecedents). Unlike the case of general
  coreference resolution, zero pronouns have to be detected prior to
  resolution because they are not expressed in discourse. Our method
  integrates two probability parameters to perform zero pronoun
  detection and resolution in a single framework. The first parameter
  quantifies the degree to which a given case is a zero pronoun. The
  second parameter quantifies the degree to which a given entity is
  the antecedent for a detected zero pronoun. To compute these
  parameters efficiently, we use corpora with/without annotations of
  anaphoric relations.  We show the effectiveness of our method by way
  of experiments.
\end{abstract}

\section{Introduction}
\label{sec:introduction}

Anaphora resolution is crucial in natural language processing (NLP),
specifically, discourse analysis. In the case of English, partially
motivated by Message Understanding Conferences
(MUCs)~\cite{grishman96}, a number of coreference resolution methods
have been proposed.

In other languages such as Japanese and Spanish, anaphoric expressions
are often omitted. Ellipses related to obligatory cases are usually
termed zero pronouns. Since zero pronouns are not expressed in
discourse, they have to be detected prior to identifying their
antecedents. Thus, although in English pleonastic pronouns have to be
determined whether or not they are anaphoric expressions prior to
resolution, the process of analyzing Japanese zero pronouns is
different from general coreference resolution in English.

For identifying anaphoric relations, existing methods are classified
into two fundamental approaches: rule-based and statistical approaches.

In rule-based
approaches~\cite{gros95,hobbs78,mitk98,naka96-2,okum96,palomar01,walk94},
anaphoric relations between anaphors and their antecedents are
identified by way of hand-crafted rules, which typically rely on
syntactic structures, gender/number agreement, and selectional
restrictions.  However, it is difficult to produce rules exhaustively,
and rules that are developed for a specific language are not
necessarily effective for other languages. For example, gender/number
agreement in English cannot be applied to Japanese.

Statistical approaches~\cite{aone95,ge98,kim95,soon01} use statistical
models produced based on corpora annotated with anaphoric relations.
However, only a few attempts have been made in corpus-based anaphora
resolution for Japanese zero pronouns. One of the reasons is that it
is costly to produce a sufficient volume of training corpora annotated
with anaphoric relations.

In addition, those above methods focused mainly on identifying
antecedents, and few attempts have been made to detect zero pronouns.

Motivated by the above background, we propose a probabilistic model
for analyzing Japanese zero pronouns combined with a detection
method. In brief, our model consists of two parameters associated with
zero pronoun detection and antecedent identification. We focus on zero
pronouns whose antecedents exist in preceding sentences to zero
pronouns because they are major referential expressions in Japanese.

Section~\ref{sec:proposed approach} explains our proposed method
(system) for analyzing  Japanese zero pronouns.
Section~\ref{sec:evaluation} evaluates our method by way of
experiments using newspaper articles.  Section~\ref{sec:related works}
discusses related research literature.

\section{A System for Analyzing Japanese Zero Pronouns}
\label{sec:proposed approach}

\subsection{Overview}
\label{sec:overview}

\begin{figure}[tb]
\begin{center}
\includegraphics[scale=0.9]{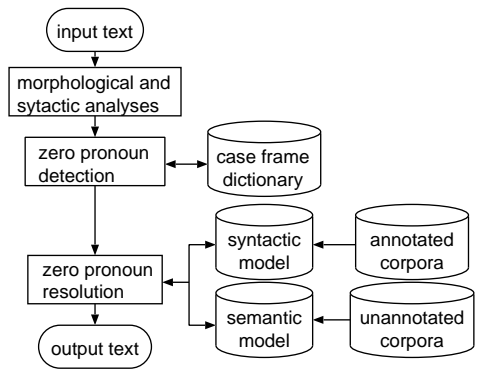}
\caption{The overall design of our system to analyze Japanese zero pronouns.}
\label{fig:overview}
\end{center}
\end{figure}

Figure~\ref{fig:overview} depicts the overall design of our system to
analyze Japanese zero pronouns. We explain the entire process based on
this figure.

First, given an input Japanese text, our system performs morphological
and syntactic analyses. In the case of Japanese, morphological
analysis involves word segmentation and part-of-speech tagging because
Japanese sentences lack lexical segmentation, for which we use the
JUMAN morphological analyzer~\cite{juman98e}. Then, we use the KNP
parser~\cite{knp98e} to identify syntactic relations between segmented
words.

Second, in a zero pronoun detection phase, the system uses syntactic
relations to detect omitted cases (nominative, accusative, and dative)
as zero pronoun candidates. To avoid zero pronouns overdetected, we
use the IPAL verb dictionary~\cite{ipal87e} including case frames
associated with 911 Japanese verbs. We discard zero pronoun candidates
unlisted in the case frames associated with a verb in question.

For verbs unlisted in the IPAL dictionary, only nominative cases are
regarded as obligatory. The system also computes a probability that
case $c$ related to target verb $v$ is a zero pronoun,
$P_{zero}(c|v)$, to select plausible zero pronoun candidates.

Ideally, in the case where a verb in question is polysemous, word
sense disambiguation is needed to select the appropriate case frame,
because different verb senses often correspond to different case
frames. However, we currently merge multiple case frames for a verb
into a single frame so as to avoid the polysemous problem.  This issue
needs to be further explored.

Third, in a zero pronoun resolution  (i.e., antecedent identification)
phase, for each zero pronoun the system extracts antecedent candidates
from the preceding contexts, which are ordered according to the extent
to which they can be the antecedent for the target zero pronoun. From
the viewpoint of probability theory, our task here is to compute a
probability that zero pronoun $\phi$ refers to antecedent $a_i$,
$P(a_i|\phi)$, and select the candidate that maximizes the probability
score. For the purpose of computing this score, we model zero pronouns
and antecedents in Section~\ref{sec:features}.

Finally, the system outputs texts containing anaphoric relations.  In
addition, the number of zero pronouns analyzed by the system can
optionally be controlled based on the certainty score described in
Section~\ref{sec:certainty}.

\subsection{Modeling Zero Pronouns and Antecedents}
\label{sec:features}

According to past literature associated with zero pronoun resolution
and our preliminary study, we use the following six features to model
zero pronouns and antecedents.

\vspace{3mm}
\noindent
$\bullet$ Features for zero pronouns
\begin{itemize}
\item[--] Verbs that govern zero pronouns ($v$), which denote verbs
  whose cases are omitted.
  
\item[--] Surface cases related to zero pronouns ($c$), for which
  possible values are Japanese case marker suffixes, {\it ga\/}
  (nominative), {\it wo\/} (accusative), and {\it ni\/} (dative).  Those
  values indicate which cases are omitted.
\end{itemize}
\noindent
$\bullet$ Features for antecedents
\begin{itemize}
\item[--] Post-positional particles ($p$), which play crucial roles in
  resolving Japanese zero pronouns~\cite{kame86,walk94}.
  
\item[--] Distance ($d$), which denotes the distance (proximity)
  between a zero pronoun and an antecedent candidate in an input text. In
  the case where they occur in the same sentence, its value takes $0$.
  In the case where an antecedent occurs in $n$ sentences previous to
  the sentence including a zero pronoun, its value takes $n$.
    
\item[--] Constraint related to relative clauses ($r$), which denotes
  whether an antecedent is included in a relative clause or not. In
  the case where it is included, the value of $r$ takes \textit{true},
  otherwise \textit{false}.  The rationale behind this feature is that
  Japanese zero pronouns tend {\em not} to refer to noun phrases in
  relative clauses.
    
\item[--] Semantic classes ($n$), which represent semantic classes
  associated with antecedents.  We use 544 semantic classes defined in
  the Japanese \textit{Bunruigoihyou} thesaurus~\cite{koku64e}, which
  contains 55,443 Japanese nouns.
    
\end{itemize}

\subsection{Our Probabilistic Model for Zero Pronoun Detection and Resolution}
\label{sec:probabilistic model}

We consider probabilities that unsatisfied case $c$ related to verb
$v$ is a zero pronoun, $P_{zero}(c|v)$, and that zero pronoun $\phi_c$
refers to antecedent $a_i$, $P(a_i|\phi_c)$. Thus, a probability that
case $c$ ($\phi_c$) is zero-pronominalized and refers to candidate
$a_i$ is formalized as in Equation~(\ref{eq:product}).
\begin{eqnarray}
  \label{eq:product}
  P(a_i|\phi_c)\cdot P_{zero}(c|v)
\end{eqnarray}
Here, $P_{zero}(c|v)$ and $P(a_i|\phi_c)$ are computed in the
detection and resolution phases, respectively (see
Figure~\ref{fig:overview}).

Since zero pronouns are omitted obligatory cases, whether or not case
$c$ is a zero pronoun depends on the extent to which case $c$ is
obligatory for verb $v$. Case $c$ is likely to be obligatory for verb
$v$ if $c$ frequently co-occurs with $v$. Thus, we compute
$P_{zero}(c|v)$ based on the co-occurrence frequency of $\langle
v,c\rangle$ pairs, which can be extracted from unannotated corpora.
$P_{zero}(c|v)$ takes 1 in the case where $c$ is $ga$ (nominative)
regardless of the target verb, because $ga$ is obligatory for most
Japanese verbs.

Given the formal representation for zero pronouns and antecedents in
Section~\ref{sec:features}, the probability, $P(a|\phi)$, is expressed
as in Equation~(\ref{eq:paz1}).
\begin{eqnarray}
  \label{eq:paz1}
  P(a_i|\phi) = P(p_i,d_i,r_i,n_i|v,c)
\end{eqnarray}

\noindent
To improve the efficiency of probability estimation, we decompose the
right-hand side of Equation~(\ref{eq:paz1}) as follows.

Since a preliminary study showed that $d_i$ and $r_i$ were relatively
independent of the other features, we approximate
Equation~(\ref{eq:paz1}) as in Equation~(\ref{eq:paz2}).
\begin{eqnarray}
  \label{eq:paz2}
  \begin{array}{@{}r@{~}c@{~}l@{}}
    \vspace*{1mm}
    P(a_i|\phi) & \approx & P(p_i,n_i|v,c)\cdot P(d_i)\cdot P(r_i)\\
    \vspace*{1mm}
    &=& P(p_i|n_i,v,c)\cdot P(n_i|v,c)\\
    && \mbox{}\cdot P(d_i)\cdot P(r_i)
  \end{array}
\end{eqnarray}
Given that $p_i$ is independent of $v$ and $n_i$, we can further
approximate Equation~(\ref{eq:paz2}) to derive
Equation~(\ref{eq:paz}).
\begin{eqnarray}
  \label{eq:paz}
  P(a_i|\phi_c) \approx P(p_i|c)\!\cdot\! P(d_i)\!\cdot\! P(r_i)\!\cdot\! P(n_i|v,c)
\end{eqnarray}
Here, the first three factors, $P(p_i|c)\cdot P(d_i)\cdot P(r_i)$, are
related to syntactic properties, and $P(n_i|v,c)$ is a semantic
property associated with zero pronouns and antecedents. We shall call
the former and latter ``syntactic'' and ``semantic'' models,
respectively.

Each parameter in Equation~(\ref{eq:paz}) is computed as in Equations
(\ref{eq:ppc}), where $F(x)$ denotes the frequency of $x$ in corpora
annotated with anaphoric relations.
\begin{eqnarray}
  \label{eq:ppc}
  \begin{array}{rcl}
    \vspace*{1mm}
    P(p_i|c)&=&\displaystyle\frac{F(p_i,c)}{\sum_{j}F(p_j,c)}\\
    \vspace*{1mm}
    \label{eq:pd}
    P(d_i)&=&\displaystyle\frac{F(d_i)}{\sum_{j}F(d_j)}\\
    \vspace*{1mm}
    \label{eq:pm}
    P(r_i)&=&\displaystyle\frac{F(r_i)}{\sum_{j}F(r_j)}\\
    \label{eq:pns}
    P(n_i|v,c)&=&\displaystyle\frac{F(n_i,v,c)}{\sum_{j}F(n_j,v,c)}
  \end{array}
\end{eqnarray}
However, since estimating a semantic model, $P(n_i|v,c)$, needs
large-scale annotated corpora, the data sparseness problem is crucial.
Thus, we explore the use of unannotated corpora.

For $P(n_i|v,c)$, $v$ and $c$ are features for a zero pronoun, and
$n_i$ is a feature for an antecedent. However, we can regard $v$, $c$,
and $n_i$ as features for a verb and its case noun because zero
pronouns are omitted case nouns. Thus, it is possible to estimate the
probability based on co-occurrences of verbs and their case nouns,
which can be extracted automatically from large-scale unannotated
corpora.

\subsection{Computing Certainty Score}
\label{sec:certainty}

Since zero pronoun analysis is not a stand-alone application, our
system is used as a module in other NLP applications, such as machine
translation. In those applications, it is desirable that erroneous
anaphoric relations are not generated. Thus, we propose a notion of
certainty to output only zero pronouns that are detected and resolved
with a high certainty score.

We formalize the certainty score, $C(\phi_c)$, for each zero pronoun
as in Equation (\ref{eq:certainty}), where $P_1(\phi_c)$ and
$P_2(\phi_c)$ denote probabilities computed by
Equation~(\ref{eq:product}) for the first and second ranked
candidates, respectively. In addition, $t$ is a parametric constant,
which is experimentally set to $0.5$.
\begin{eqnarray}
  \label{eq:certainty}
  C(\phi_c) = t\!\cdot\! P_1(\phi_c)+(1\!-\!t)(P_1(\phi_c)\!-\!P_2(\phi_c))
\end{eqnarray}
The certainty score becomes great in the case where $P_1(\phi_c)$ is
sufficiently great and significantly greater than $P_2(\phi_c)$.

\section{Evaluation}
\label{sec:evaluation}

\subsection{Methodology}
\label{sec:methodology}

To investigate the performance of our system, we used
\textit{Kyotodaigaku} Text Corpus version 2.0~\cite{kuro98}, in which
20,000 articles in \textit{Mainichi Shimbun} newspaper articles in
1995 were analyzed by JUMAN and KNP (i.e., the morph/syntax analyzers
used in our system) and revised manually. From this corpus, we
randomly selected 30 general articles (e.g., politics and sports) and
manually annotated those articles with anaphoric relations for zero
pronouns. The number of zero pronouns contained in those articles was
449.

We used a leave-one-out cross-validation evaluation method: we
conducted 30 trials in each of which  one article was used as a test
input and the remaining 29 articles were used for producing a
syntactic model. We used six years worth of \textit{Mainichi Shimbun}
newspaper articles~\cite{mainichi-e} to produce a semantic model based
on co-occurrences of verbs and their case nouns.

To extract verbs and their case noun pairs from newspaper articles, we
performed a morphological analysis by JUMAN and extracted dependency
relations using a relatively simple rule: we assumed that each noun
modifies the verb of highest proximity.  As a result, we obtained 12
million co-occurrences associated with 6,194 verb types. Then, we
generalized the extracted nouns into semantic classes in the Japanese
{\it Bunruigoihyou\/} thesaurus. In the case where a noun was
associated with multiple classes, the noun was assigned to all
possible classes.
In the case where a noun was not listed in the thesaurus, the noun
itself was regarded as a single semantic class.

\begin{table*}[htb]
  \begin{center}
    \caption{Experimental results for zero pronoun resolution.}
    \label{tab:riyousosei}
    \footnotesize
    \smallskip
    \begin{tabular}{cr@{~}lccr@{~}lcc} \hline\hline
      & \multicolumn{8}{c}{\# of Correct cases (Accuracy)} \\
      \cline{2-9}
      $k$ & \multicolumn{2}{c}{$Sem1$} & $Sem2$ & $Syn$ & \multicolumn{2}{c}{$Both1$} & $Both2$ & $Rule$\\
      \hline
      1 & 25 & (6.2\%)  & 119 (29.5\%) & 185 (45.8\%) & 30 & (7.4\%) & \bf{205 (50.7\%)} & 162 (40.1\%) \\
      2 & 46 & (11.4\%) & 193 (47.8\%) & 227 (56.2\%) & 49 & (12.1\%) & \bf{250 (61.9\%)} & 213 (52.7\%) \\
      3 & 72 & (17.8\%) & 230 (56.9\%) & 262 (64.9\%) & 75 & (18.6\%) & \bf{280 (69.3\%)} & 237 (58.6\%) \\
      \hline
    \end{tabular}
  \end{center}
\end{table*}

\subsection{Comparative Experiments}
\label{sec:results}

Fundamentally, our evaluation is two-fold: we evaluated only zero
pronoun resolution (antecedent identification) and a combination of
detection and resolution. In the former case, we assumed that all the
zero pronouns are correctly detected, and investigated the
effectiveness of the resolution model, $P(a_i|\phi)$. In the latter
case, we investigated the effectiveness of the combined model,
$P(a_i|\phi_c)\cdot P_{zero}(c|v)$.

First, we compared the performance of the following different models
for zero pronoun resolution, $P(a_i|\phi)$:
\begin{list}{$\bullet$}{\itemsep 3pt \parsep 0pt}
\item a semantic model produced based on annotated corpora ($Sem1$),
\item a semantic model produced based on unannotated corpora, using
  co-occurrences of verbs and their case nouns ($Sem2$),
\item a syntactic model ($Syn$),
\item a combination of $Syn$ and $Sem1$ ($Both1$),
\item a combination of $Syn$ and $Sem2$ ($Both2$), which is our
  complete model for zero pronoun resolution,
\item a rule-based model ($Rule$).
\end{list}
As a control (baseline) model, we took approximately two man-months to
develop a rule-based model ($Rule$) through an analysis on ten
articles in \textit{Kyotodaigaku} Text Corpus. This model uses rules
typically used in existing rule-based methods: 1) post-positional
particles that follow antecedent candidates, 2) proximity between zero
pronouns and antecedent candidates, and 3) conjunctive particles. We
did not use semantic properties in the rule-based method because they
decreased the system accuracy in a preliminary study.

Table~\ref{tab:riyousosei} shows the results, where we regarded the
$k$-best antecedent candidates as the final output and compared
results for different values of $k$.  In the case where the correct
answer was included in the $k$-best candidates, we judged it
correct. In addition, ``Accuracy'' is the ratio between the number of
zero pronouns whose antecedents were correctly identified and the
number of zero pronouns correctly detected by the system (404 for all
the models). Bold figures denote the highest performance for each
value of $k$ across different models. Here, the average number of
antecedent candidates per zero pronoun was 27 regardless of the model,
and thus the accuracy was 3.7\% in the case where the system randomly
selected antecedents.

Looking at the results for two different semantic models, $Sem2$
outperformed $Sem1$, which indicates that the use of co-occurrences of
verbs and their case nouns was effective to identify antecedents and
avoid the data sparseness problem in producing a semantic model.

The syntactic model, $Syn$, outperformed the two semantic models
independently, and therefore the syntactic features used in our model
were more effective than the semantic features to identify
antecedents. When both syntactic and semantic models were used in
$Both2$, the accuracy was further improved.  While the rule-based
method, $Rule$, achieved a relatively high accuracy, our complete
model, $Both2$, outperformed $Rule$ irrespective of the value of $k$.
To sum up, we conclude that both syntactic and semantic models were
effective to identify appropriate anaphoric relations.

At the same time, since our method requires annotated corpora, the
relation between the corpus size and accuracy is crucial. Thus, we
performed two additional experiments associated with $Both2$.

\begin{figure}[tb]
\begin{center}
  \includegraphics[scale=0.62]{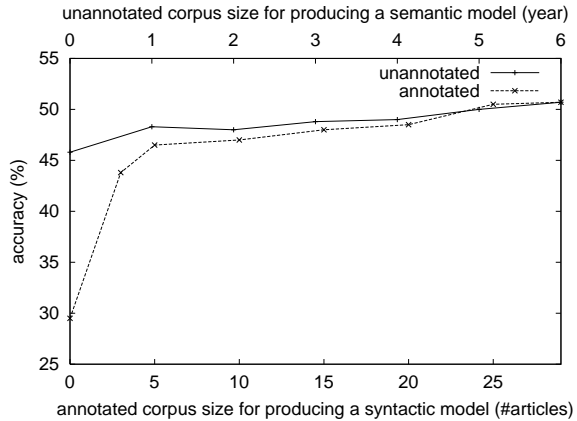}
\caption{The relation between the corpus size and accuracy for a combination of syntactic and semantic models ($Both2$).}
\label{fig:both}
\end{center}
\end{figure}

In the first experiment, we varied the number of annotated articles
used to produce a syntactic model, where a semantic model was produced
based on six years worth of newspaper articles. In the second
experiment, we varied the number of unannotated articles used to
produce a semantic model, where a syntactic model was produced based
on 29 annotated articles. In Figure~\ref{fig:both}, we show two {\em
  independent\/} results as space is limited: the dashed and solid
graphs correspond to the results of the first and second experiments,
respectively. Given all the articles for modeling, the resultant
accuracy for each experiment was 50.7\%, which corresponds to that for
$Both2$ with $k=1$ in Table~\ref{tab:riyousosei}.

In the case where the number of articles was varied in producing a
syntactic model, the accuracy improved rapidly in the first five
articles. This indicates that a high accuracy can be obtained by a
relatively small number of supervised articles.  In the case where the
amount of unannotated corpora was varied in producing a semantic
model, the accuracy marginally improved as the corpus size
increases. However, note that we do not need human supervision to
produce a semantic model.

\begin{figure}
\begin{center}
\includegraphics[scale=0.60]{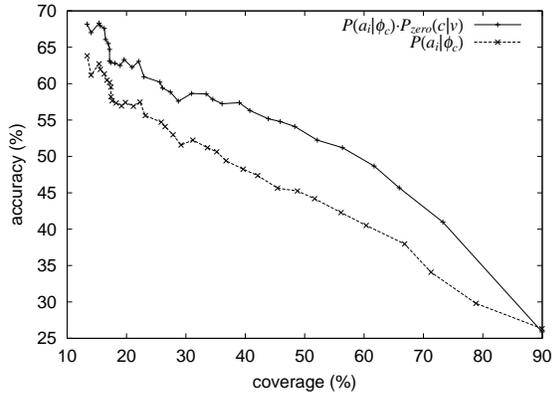}
\caption{The relation between coverage and accuracy for zero pronoun detection ({\it Both\/}2).}
\label{fig:detection}
\end{center}
\end{figure}

\begin{figure}
\begin{center}
  \includegraphics[scale=0.60]{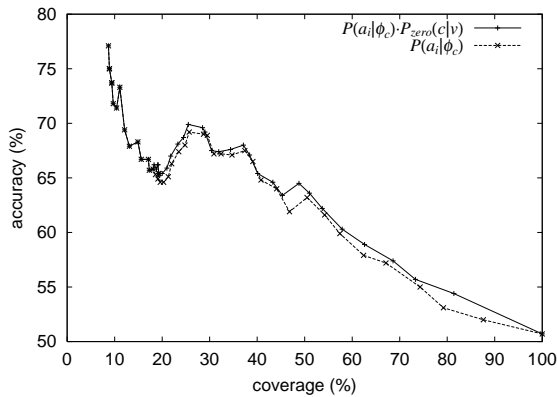}
\caption{The relation between coverage and accuracy for antecedent identification ({\it Both\/}2).}
\label{fig:sikii}
\end{center}
\end{figure}

Finally, we evaluated the effectiveness of the combination of zero
pronoun detection and resolution in Equation~(\ref{eq:product}).  To
investigate the contribution of the detection model, $P_{zero}(c|v)$,
we used $P(a_i|\phi_c)$ for comparison. Both cases used $Both2$ to
compute the probability for zero pronoun resolution.  We varied a
threshold for the certainty score to plot coverage-accuracy graphs for
zero pronoun detection (Figure~\ref{fig:detection}) and antecedent
identification (Figure~\ref{fig:sikii}).

In Figure~\ref{fig:detection}, ``coverage'' is the ratio between the
number of zero pronouns correctly detected by the system and the total
number of zero pronouns in input texts, and ``accuracy'' is the ratio
between the number of zero pronouns correctly detected and the total
number of zero pronouns detected by the system. Note that since our
system failed to detect a number of zero pronouns, the coverage could
not be 100\%.

Figure~\ref{fig:detection} shows that  as the coverage
decreases, the accuracy improved irrespective of the model used.  When
compared with the case of $P(a_i|\phi)$, our model, $P(a_i|\phi)\cdot
P_{zero}(c|v)$, achieved a higher accuracy regardless of the coverage.

In Figure~\ref{fig:sikii}, ``coverage'' is the ratio between the
number of zero pronouns whose antecedents were generated and the
number of zero pronouns correctly detected by the system.  The
accuracy was improved by decreasing the coverage, and our model
marginally improved the accuracy for $P(a_i|\phi)$.

According to those above results, our model was effective to improve
the accuracy for zero pronoun detection and did not have side effect
on the antecedent identification process. As a result, the overall
accuracy of zero pronoun detection and resolution was improved.

\section{Related Work}
\label{sec:related works}

Kim and Ehara~\shortcite{kim95} proposed a probabilistic model to
resolve subjective zero pronouns for the purpose of Japanese/English
machine translation. In their model, the search scope for possible
antecedents was limited to the sentence containing zero pronouns.  In
contrast, our method can resolve zero pronouns in both
intra/inter-sentential anaphora types.

Aone and Bennett~\shortcite{aone95} used a decision tree to determine
appropriate antecedents for zero pronouns. They focused on proper and
definite nouns used in anaphoric expressions as well as zero pronouns.
However, their method resolves only anaphors that refer to
organization names (e.g., private companies), which are generally
easier to resolve than our case.

Both above existing methods require annotated corpora for statistical
modeling, while we used corpora with/without annotations related to
anaphoric relations, and thus we can easily obtain large-scale corpora
to avoid the data sparseness problem.

Nakaiwa~\shortcite{nakaiwa00} used Japanese/English bilingual corpora
to identify anaphoric relations of Japanese zero pronouns by comparing
J/E sentence pairs. The rationale behind this method is that
obligatory cases zero-pronominalized in Japanese are usually expressed
in English. However, in the case where corresponding English
expressions are pronouns and anaphors, their method is not
effective. Additionally, bilingual corpora are more expensive to
obtain than monolingual corpora used in our method.

Finally, our method integrates a parameter for zero pronoun detection
in computing the certainty score. Thus, we can improve the accuracy of
our system by discarding extraneous outputs with a small certainty
score.

\section{Conclusion}
\label{sec:conclusion}

We proposed a probabilistic model to analyze Japanese zero pronouns
that refer to antecedents in the previous context. Our model consists
of two probabilistic parameters corresponding to detecting zero
pronouns and identifying their antecedents, respectively. The latter
is decomposed into syntactic and semantic properties. To estimate
those parameters efficiently, we used annotated/unannotated
corpora. In addition, we formalized the certainty score to improve the
accuracy. Through experiments, we showed that the use of unannotated
corpora was effective to avoid the data sparseness problem and that
the certainty score further improved the accuracy.

Future work would include word sense disambiguation for polysemous
predicate verbs to select appropriate case frames in the zero pronoun
detection process.

\bibliographystyle{acl}

\end{document}